\documentclass[11pt,a4paper]{article}
\usepackage[hyperref]{emnlp2020}
\usepackage{times}
\usepackage{latexsym}

\usepackage{graphicx}
\usepackage{CJKutf8}
\usepackage{multirow}
\usepackage{array}
\usepackage{tabularx}
\usepackage{makecell}
\usepackage{amsmath} 
\usepackage{amssymb}
\usepackage{tablefootnote}
\usepackage{subcaption}
\usepackage{diagbox}
\usepackage{algorithm}
\usepackage[noend]{algpseudocode}
\usepackage{booktabs}
\usepackage{xcolor}
\usepackage{color, soul} 
\usepackage{colortbl}
\usepackage{makecell}

\usepackage{microtype}

\newcolumntype{L}[1]{>{\raggedright\arraybackslash}p{#1}}
\newcolumntype{C}[1]{>{\centering\arraybackslash}p{#1}}
\newcolumntype{R}[1]{>{\raggedleft\arraybackslash}p{#1}}

\DeclareSymbolFont{extraup}{U}{zavm}{m}{n}
\DeclareMathSymbol{\varheart}{\mathalpha}{extraup}{86}
\DeclareMathSymbol{\vardiamond}{\mathalpha}{extraup}{87}

\usepackage{microtype}

\aclfinalcopy 
\def\aclpaperid{1419} 

\title{Improving Constituency Parsing with Span Attention}

\author{
    Yuanhe Tian$^{\varheart}$, \hspace{0.2cm}
    Yan Song$^{{\spadesuit}\heartsuit\dag}$, \hspace{0.2cm}
    Fei Xia$^{\varheart}$, \hspace{0.2cm}
    Tong Zhang$^{\vardiamond}$ \\
    $^{\varheart}$University of Washington \hspace{0.4cm}
    $^{\spadesuit}$The Chinese University of Hong Kong (Shenzhen)\\
    $^{\heartsuit}$Shenzhen Research Institute of Big Data \\
    $^{\vardiamond}$The Hong Kong University of Science and Technology\\
    $^{\varheart}$\texttt{\{yhtian, fxia\}@uw.edu} \hspace{0.4cm} 
    $^{\spadesuit}$\texttt{songyan@cuhk.edu.cn} \\
    $^{\vardiamond}$\texttt{tongzhang@ust.hk} \hspace{0.4cm}
}

\date{}

\begin{document}
\maketitle

\renewcommand{\thefootnote}{\fnsymbol{footnote}}
\footnotetext[2]{Corresponding author.}

\renewcommand{\thefootnote}{\arabic{footnote}}

\begin{abstract}

Constituency parsing is a fundamental and important task for natural language understanding, 
where a good representation of contextual information can help this task.
N-grams, which is a conventional type of feature for contextual information, have been demonstrated to be useful in many tasks, and thus could also be beneficial for constituency parsing if they are appropriately modeled.
%
In this paper, we propose span attention for neural chart-based constituency parsing to leverage n-gram information.
Considering that current chart-based parsers with Transformer-based encoder represent spans by subtraction of the hidden states at the span boundaries, which may cause information loss especially for long spans, we incorporate n-grams into span representations by weighting them according to their contributions to the parsing process.
Moreover, we propose categorical span attention to further enhance the model
by weighting n-grams within different length categories,
and thus benefit long-sentence parsing.
Experimental results on three widely used benchmark datasets demonstrate the effectiveness of our approach in parsing 
Arabic, Chinese, and English, where state-of-the-art performance is obtained by our approach on all of them.\footnote{Our code and the best performing models are released at \url{https://github.com/cuhksz-nlp/SAPar}.}

\end{abstract}

\section{Introduction}
\label{intro}

Constituency parsing, which aims to generate a structured syntactic parse tree for a given sentence, is one of the most fundamental tasks in natural language processing (NLP), and plays an important role in many downstream tasks such as relation extraction \cite{jiang-diesner-2019-constituency}, natural language inference \cite{chen-etal-2017-enhanced}, and machine translation \cite{ma-etal-2018-forest}.
Recently, neural parsers \cite{NIPS2015_5635,dyer-etal-2016-recurrent,stern-etal-2017-minimal,kitaev-etal-2019-multilingual} without using any grammar rules significantly outperform conventional statistical grammar-based ones \cite{collins-1997-three,sagae-lavie-2005-classifier,glaysher-moldovan-2006-speeding,song2009pcfg}, because neural networks, especially recurrent models (e.g, Bi-LSTM), are adept in capturing long range contextual information, which is essential to modeling the entire sentence.
Particularly, a significant boost on the performance of chart-based parsers is observed from some recent studies \cite{kitaev-klein-2018-constituency,kitaev-etal-2019-multilingual,zhou-zhao-2019-head} that employ advanced text encoders (i.e., Transformer, BERT, and XLNet), which further demonstrates the usefulness of contexts for parsing.
%

%
\begin{figure}
    \centering
    \includegraphics[width=0.45\textwidth, trim=0 15 0 10]{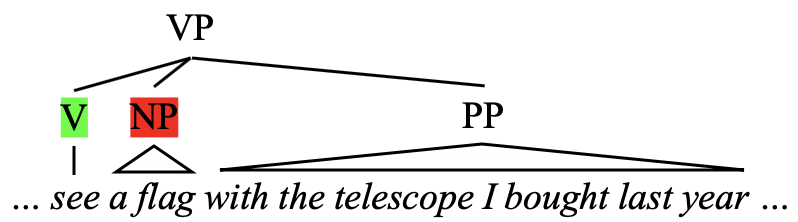}
    \caption{
    The treelet of an example of the form ``\textit{V}+\textit{NP}+\textit{PP}", where the ``\textit{PP}'' should attach to the ``\textit{V}'' (in green) rather than the ``NP'' (in red).}
    \label{fig: intro_example}
    \vskip -1.5em
\end{figure}


\begin{figure*}[t]
    \centering
    \includegraphics[width=1.0\textwidth, trim=0 18 0 0]{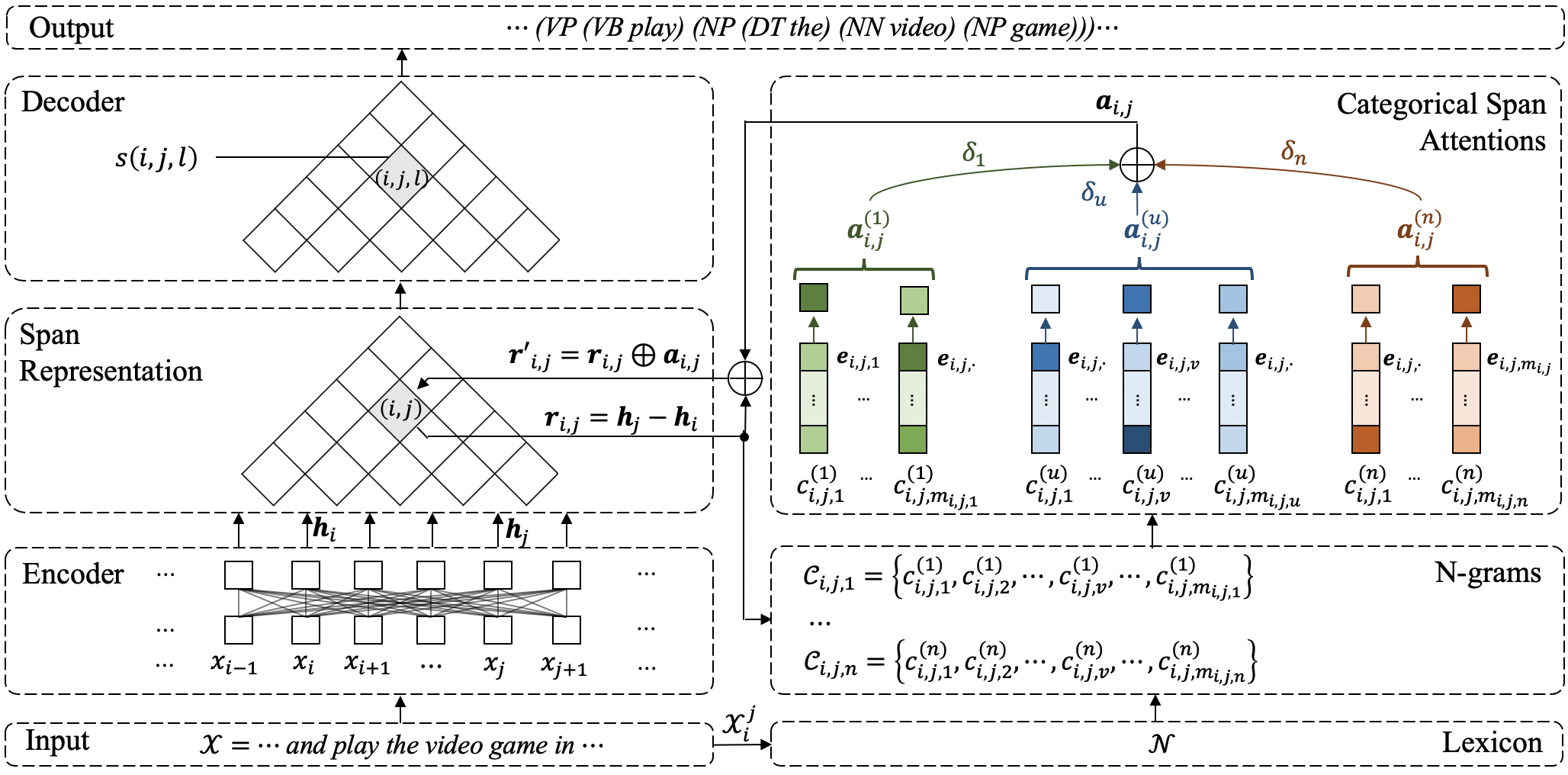}
    \caption{
    The architecture of the chart-based constituency parser with span attention, 
    with an example partial input sentence and its output.
    The right part of the figure shows the categorical span attention, where extracted n-grams in span $(i,j)$ are categorized by their length so that n-grams in different categories are weighted separately (different colors refer to different n-gram categories).
    Note that for normal span attention, all n-grams are weighted together, where attention $\mathbf{a}_{i,j}$ directly corresponds to $\mathbf{e}_{i,j,\cdot}$ in the figure.}
    \label{fig:model}
    \vskip -1em
\end{figure*}

In general, besides powerful encoders, other extra information (such as pre-trained embeddings and extra syntactic information) can also provide useful contextual information and thus enhance model performance in many NLP tasks \cite{pennington2014glove,song-etal-2018-joint,zhang-etal-2019-incorporating,mrini2019rethinking,tian-etal-2020-joint,tian-etal-2020-suppertagging}.
As one type of the extra information, n-grams are used as a simple yet effective source of contextual feature in many studies \cite{song-etal-2009-transliteration,song2012using,yoon-etal-2018-learning,tian2020improving}
%
Therefore, they could be potentially beneficial for parsing as well.
However, recent chart-based parers \cite{stern-etal-2017-minimal,kitaev-klein-2018-constituency,gaddy-etal-2018-whats,kitaev-etal-2019-multilingual,zhou-zhao-2019-head} make rare effort to leverage such n-gram information.
Another potential issue with current chart-based parsers is that they represent spans by subtraction of hidden states at the span boundaries, where the context information in-between may be lost and thus hurt parsing performance especially for long sentences.
N-grams can be a simple yet useful source to fill the missing information.
%
For instance, Figure \ref{fig: intro_example} illustrates the treelet of an example in the form of  ``\textit{V}+\textit{NP}+\textit{PP}''. As a classic example of PP-attachment ambiguity, a parser may wrongly attach the ``\textit{PP}'' to the ``\textit{NP}'' if it only focuses on the words at the boundaries of the text span ``\textit{flag ... year}'' and in-between information is not represented properly.
In this case, n-grams within that span (e.g., the uni-gram ``\textit{telescope}'') can provide useful cues indicating that the ``\textit{PP}'' should be attached to the ``\textit{V}''.
Although there are traditional non-neural parsers using n-grams as features to improve parsing \cite{sagae-lavie-2005-classifier,pitler-etal-2010-using}, they are limited in treating them euqally without learning their weights.
Therefore, unimportant n-grams may deliver misleading information and lead to wrong predictions.

To address this problem, in this paper, we propose a span attention module to enhance chart-based neural constituency parsing by incorporating appropriate n-grams into span representations.
%
Specifically, for each text span we extract all its substrings that appear in an n-gram lexicon; the span attention uses the normal attention mechanism to weight them with respect to their contributions to predict the constituency label of the span.
Because in general short n-grams occur more frequently than long ones, they may dominate in the attention if all n-grams are globally weighted,
We further enhance our approach with a categorical mechanism which first groups n-grams into different categories according to their length and then weights them within each category.
Thus, n-grams with different lengths are separately treated and the infrequent long ones carrying more contextual information can be better leveraged.
%
%
The effectiveness of our approach is illustrated by experimental results on three benchmark datasets from different languages (i.e., Arabic, Chinese, and English), on all of which state-of-the-art performance is achieved.

\section{The Approach}

Our approach follows the chart-based paradigm for constituency parsing, where the parse tree $\mathcal{T}$ of an input sentence $\mathcal{X}=x_{1}x_{2} \cdots x_{i} \cdots x_{j} \cdots x_{q}$ is represented as a set of labeled spans. A span is denoted by a triplet $(i, j, l)$ with $i$ and $j$ referring to the beginning and ending positions of a span with a label $l \in \mathcal{L}$.
Here, $\mathcal{L}$ is the label set containing $d_{l}$ constituent types.
The architecture of our approach is shown in Figure \ref{fig:model}. The left side is the backbone chart-based parser.
It assigns real value scores $s(i, j, l)$ to the labeled spans, then computes the score of a candidate tree by summing up the scores of all its spans, and finally chooses a valid tree $\widehat{\mathcal{T}}$ with the highest score $s$ by
\begin{align}
\setlength\abovedisplayskip{5pt}
\setlength\belowdisplayskip{5pt}
    \widehat{\mathcal{T}} = \underset{\mathcal{T}}{\arg \max} \sum_{\begin{subarray}{c}(i,j,l) \in \mathcal{T} \\ 0 < i < j \leq q\end{subarray}} s(i,j,l)
\end{align}

The right side of Figure \ref{fig:model} shows the proposed span attention to enhance the backbone parser,
where n-grams in $\mathcal{X}$ are extracted from a pre-constructed lexicon $\mathcal{N}$ and are weighted through the attention module according to their contribution to the parsing process.
Therefore, the process of computing $s(i, j, l)$ of the labeled spans through our approach is formalized by
\begin{align}
\setlength\abovedisplayskip{5pt}
\setlength\belowdisplayskip{5pt}
    s(i, j, l) = 
    p(l|\mathcal{X}, \mathcal{SA}(\mathcal{X}_{i}^{j}, \mathcal{N}))
\end{align}
where $\mathcal{X}_{i}^{j}$ is the text in range $[i,j]$ of $\mathcal{X}$; $\mathcal{SA}$ represents the span attention module and $p$ computes the probability of labeling $l \in \mathcal{L}$ to the span $(i,j)$.

In this section,
we start with a brief introduction of neural chart-based parsing, then describe our span attention, and 
end with an illustration of incorporating span attention into the parsing process.

\subsection{Neural Chart-based Parsing}
\label{sec: neural chart parser}

Recent neural chart-based parsers \cite{stern-etal-2017-minimal,kitaev-klein-2018-constituency,kitaev-etal-2019-multilingual,zhou-zhao-2019-head} follow the encoder-decoder way, where the encoder receives $\mathcal{X}$ and generates a sequence of context-sensitive hidden vectors (denoted as $\mathbf{h}_{i}$ and $\mathbf{h}_{j}$ for $x_{i}$ and $x_{j}$, respectively), which are used to compute the span representation $\mathbf{r}_{i,j} \in \mathbb{R}^{d_{r}}$ for $(i,j)$
by subtraction: $\mathbf{r}_{i,j} = \mathbf{h}_{j} - \mathbf{h}_{i}$.
This span representation assumes that, for a recurrent model, e.g., LSTM, its hidden vector at each time step relies on the previous ones so that such subtraction could, to some extent, capture the contextual information of all the words in that span.\footnote{Note that this paper focuses on improving the current best performing span representation (i.e., by hidden vector subtraction) proposed by \citet{stern-etal-2017-minimal} so as to make a fair comparison, although there are other possible approaches to representing a span (e.g., max pooling).}

For decoders, most recent neural chart-based parsers follow the strategy proposed by \newcite{stern-etal-2017-minimal}, where all span representations $\mathbf{r}_{i,j}$ are fed into a variant of CYK algorithm to generate a globally optimized tree for each sentence.
Normally, $\mathbf{r}_{i,j}$ is fed into multi-layer perceptrons (MLP) to compute its scores $s(i,j,\cdot)$ over the label set $\mathcal{L}$.
Afterwards, a recursion function is applied to find the highest score $s^{*}(i,j)$ of span $(i,j)$, which is computed by searching the best constituency label and the corresponding boundary $k$ $(i < k < j)$ by
\begin{equation} \label{eq: best (i,j)}
\setlength\abovedisplayskip{5pt}
\setlength\belowdisplayskip{5pt}
    \begin{array}{ll}
        s^{*}(i,j) &= \max\limits_{l \in \mathcal{L}} s(i,j,l) \\
        &+ \max\limits_{i < k < j}[s^{*}(i,k) + s^{*}(k,j)]
    \end{array}
\end{equation}
Note that in the special case where $j=i+1$, the best score only relies on the candidate label:
\begin{align} \label{eq: best (i,i+1)}
\setlength\abovedisplayskip{5pt}
\setlength\belowdisplayskip{5pt}
    s^{*}(i,j) = \max\limits_{l \in \mathcal{L}} s(i,j,l)
\end{align}
Therefore, to parse the entire sentence, one computes $s^*(1, q)$ through the above steps and use a back pointer to recover the full tree structure.

\subsection{Span Attention}

Although the encoding from subtraction of hidden states is demonstrated to be effective \cite{stern-etal-2017-minimal,kitaev-klein-2018-constituency,kitaev-etal-2019-multilingual}, 
the subtraction might not represent all the crucial information in the text span.
Especially, for Transformer-based encoders, unlike recurrent models, their $\mathbf{h}_{i}$ and $\mathbf{h}_{j}$ have no strong dependency on each other so that subtraction may fail to fully capture the contextual information in the span, especially when the span is long.
%
Since n-grams are a good source of the information in the text span, we propose span attention to incorporate weighted n-gram information into span representations to help score the spans $(i,j,l)$.

In detail, for each span $(i,j)$ in $\mathcal{X}$, we extract all n-grams in that span that appear in Lexicon $\mathcal{N}$ to form a set $\mathcal{C}_{i,j} = \{c_{i,j,1}, c_{i,j,2}, \cdots c_{i,j,v}, \cdots c_{i,j, m_{i,j}}\}$ and use the set in span attention.
The attention of each n-gram $c_{i,j,v}$ for $(i,j)$ is activated by
\begin{equation} \label{eq: a_ijv}
\setlength\abovedisplayskip{5pt}
\setlength\belowdisplayskip{5pt}
    a_{i,j,v} = \frac{\exp(\mathbf{r}_{i,j}^{\top} \cdot \mathbf{e}_{i,j,v})}
                    {\sum_{v=1}^{m_{i,j}} \exp(\mathbf{r}_{i,j}^{\top} \cdot \mathbf{e}_{i,j,v})}
\end{equation}
where $\mathbf{e}_{i,j,v} \in \mathbb{R}^{d_{r}}$ is the embedding of $c_{i,j,v}$ whose dimension is identical to that of $\mathbf{r}_{i,j}$.
The resulted attention vector $\mathbf{a}_{i,j} \in \mathbb{R}^{d_{r}}$ is thus computed by the weighted average of n-gram embeddings by
\begin{equation}
\label{eq: a_ij}
\setlength\abovedisplayskip{5pt}
\setlength\belowdisplayskip{5pt}
    \mathbf{a}_{i,j} = \sum_{v=1}^{m_{i,j}} a_{i,j,v} \mathbf{e}_{i,j,v}
\end{equation}
and it is used to enhance the span representation.

In normal attention, all n-grams are weighted globally and short n-grams may dominate the attention because they occur much more frequently than long ones and are intensively updated.
However, there are cases that long n-grams can play an important role in parsing when they carry useful context and boundary information.
%
Therefore, we extend the span attention with a category mechanism (namely, categorical span attention) by grouping n-grams based on their lengths and weighting them within each category.\footnote{We use length as the categorization criterion because (1) n-gram frequencies vary in different datasets and it is hard to find an appropriate scheme to divide them;
(2) n-grams with the same length may have similar ability to deliver contextual information so they are suitable to be grouped by such ability.}
In doing so, all n-grams in $\mathcal{N}$ are categorized into $n$ groups according to their lengths, i.e., $\mathcal{C}_{i,j}=\{\mathcal{C}_{i,j,1}, \mathcal{C}_{i,j,2}, \cdots \mathcal{C}_{i,j,u}, \cdots \mathcal{C}_{i,j,n}\}$, with $u \in [1, n]$ denoting the n-gram length.
Then, for each category with n-grams in length $u$, we follow 
the same process in Eq. (\ref{eq: a_ijv}) and (\ref{eq: a_ij}) to compute $a^{(u)}_{i,j,v}$ and $\mathbf{a}^{(u)}_{i,j}$.
The final attention is obtained from the concatenation of all categorical attentions by
\begin{equation}
\label{eq: a_iju}
\setlength\abovedisplayskip{5pt}
\setlength\belowdisplayskip{5pt}
\mathbf{a}_{i,j} = \bigoplus_{1\leq u\leq n} \delta_{u} \mathbf{a}^{(u)}_{i,j} 
\end{equation}
with a trainable parameter $\delta_{u} \in \mathbb{R}^{+}$ to balance the contribution of attentions from different categories.

\subsection{Parsing with Span Attention}

The backbone parser follows \citet{kitaev-etal-2019-multilingual} to use BERT as the encoder, where $\mathbf{r}_{i,j}=\mathbf{h}_{j}-\mathbf{h}_{i}$ is applied to represent the span $(i,j)$.
Once $\mathbf{a}_{i,j}$ is obtained from the span attention for $(i,j)$, we incorporate it into the backbone parsing process by directly concatenating it with $\mathbf{r}_{i,j}$:
$\mathbf{r}'_{i,j}=\mathbf{r}_{i,j} \oplus \mathbf{a}_{i,j} \in \mathbb{R}^{d_{r} \cdot (n+1)}$.
Then, we apply two fully connected layers with $ReLU$ activation function to $\mathbf{r}'_{i,j}$ and compute the span scores $s(i,j,\cdot)$ over the label set $\mathcal{L}$, which can be formalized by:
\begin{align} \label{eq: o_ij}
\setlength\abovedisplayskip{5pt}
\setlength\belowdisplayskip{5pt}
        \mathbf{o}_{i,j} &= ReLU(LN(\mathbf{W}_{1} \cdot \mathbf{r}'_{i,j} + \mathbf{b}_{1}))
\end{align}
and
\begin{align}
\setlength\abovedisplayskip{5pt}
\setlength\belowdisplayskip{5pt}
    s(i,j, \cdot) &= \mathbf{W}_{2} \cdot \mathbf{o}_{i,j} + \mathbf{b}_{2}
\end{align}
Here, $LN$ denotes the layer normalization operation;
$\mathbf{W}_{1}$, $\mathbf{W}_{2}$ and $\mathbf{b}_{1}$, $\mathbf{b}_{2}$ are trainable parameters in the fully connected layers.
%
Afterwards, we use Eq. (\ref{eq: best (i,j)}) and (\ref{eq: best (i,i+1)}) to recursively find the highest score $s_{best}(1,q)$, and use a back pointer to recover the globally optimized parse tree.

\section{Experimental Settings}

\subsection{Datasets}

We test our approach on Arabic, Chinese and English benchmark datasets, namely part 1-3 of the Arabic Penn Treebank 2.0 (ATB) \cite{maamouri2004penn}, the Chinese Penn Treebank 5 (CTB5) \cite{xue2005penn}, and 
Penn Treebank 3 (PTB) \cite{marcus-etal-1993-building}.\footnote{All the datasets are obtained from the official release of Linguistic Data Consortium. The catalog numbers for ATB part 1-3 are LDC2003T06, LDC2004T02, LDC2005T20, for CTB5 is LDC2005T01, and for PTB is LDC99T42.}
For ATB, we follow \newcite{chiang-etal-2006-parsing} and \newcite{green-manning-2010-better} to use their split\footnote{Such split uses the ``Johns Hopkins 2005 Workshop'' standard, for which we follow the detailed split guideline offered by \url{https://nlp.stanford.edu/software/parser-arabic-data-splits.shtml}.} to get the training/dev/test sets and convert the texts in the dataset from Buckwalter transliteration\footnote{\url{http://languagelog.ldc.upenn.edu/myl/ldc/morph/buckwalter.html}} to modern standard Arabic.
For CTB5 and PTB, we follow \newcite{shen-etal-2018-straight} and \newcite{kamigaito-etal-2017-supervised} to split the datasets.
%
Moreover, we use the Brown Corpus \cite{marcus-etal-1993-building} and Genia \cite{tateisi-etal-2005-syntax} for cross-domain experiments.\footnote{The Brown Corpus is obtained together with PTB (LDC99T42), and the Genia corpus is obtained by its official PTB format from \url{https://nlp.stanford.edu/~mcclosky/biomedical.html}.}
For all datasets, we follow \newcite{suzuki-etal-2018-empirical} to clean up the raw data\footnote{We use the clean-up code from \url{https://github.com/nikitakit/parser-data-gen}.} and report the statistics of each resulted dataset in Table \ref{tab: dataset details}.

\begin{table}[t]
\begin{center}
\begin{small}
\begin{sc}
\begin{tabular}{L{1.2cm} | L{1.05cm} | R{1.1cm} R{1.2cm} | R{1.0cm}}
    \toprule
    \multicolumn{2}{c|}{\textbf{Datasets}} &
    \textbf{Sent} & \textbf{Token} & \textbf{ASL}\\
    \midrule
    \multirow{3}{*}{ATB} & 
    Train & 16K & 596K & 31.4 \\
    &
    Dev & 2K & 70K & 30.5 \\
    &
    Test & 2K & 70K & 29.9 \\
    \midrule
    \multirow{3}{*}{CTB5} & 
    Train & 17K & 478K & 27.4 \\
    &
    Dev & 350 & 7K & 19.5  \\
    &
    Test & 348 & 8K & 23.0 \\
    \midrule
    \multirow{3}{*}{PTB} & 
    Train & 40K & 950K & 23.9 \\
    &
    Dev & 2K & 40K & 23.6 \\
    &
    Test & 2K & 57K & 23.5 \\
    \midrule
    \multicolumn{2}{l|}{Brown (Full)} & 24K & 458K & 19.0 \\
    \addlinespace[0.04cm]
    \multicolumn{2}{l|}{Genia (Full)} & 17K & 446K & 26.2 \\
    \bottomrule
\end{tabular}
\end{sc}
\end{small}
\vspace{-0.5cm}
\end{center}
\caption{The statistics of all experimental datasets (with splits) in terms of sentence and token numbers, and average sentence length (ASL).
}
\label{tab: dataset details}
\vskip -1.4em
\end{table}

\begin{table*}[t]
\begin{center}
\begin{small}
\begin{sc}
\begin{tabular}{C{0.8cm} | L{1.7cm} | L{0.85cm} |
            C{0.82cm} C{0.82cm} >{\columncolor[rgb]{0.95,0.95,0.95}}C{0.82cm}
            | C{0.83cm} ||
            L{0.85cm} | C{0.82cm} C{0.82cm} >{\columncolor[rgb]{0.95,0.95,0.95}}C{0.82cm}
            | C{0.82cm} }
    \toprule
    \multirow{2}{*}{\textbf{Data}} & \multirow{2}{*}{\textbf{Models}} & \multicolumn{5}{c||}{\textbf{- POS}} & \multicolumn{5}{c}{\textbf{+ POS}}\\
    \addlinespace[0.02cm]
    \cline{3-12}
    \addlinespace[0.02cm]
    & 
    & \textbf{Parm} & \textbf{P} & \textbf{R} & \textbf{F1} & \textbf{M}
    & \textbf{Parm} & \textbf{P} & \textbf{R} & \textbf{F1} & \textbf{M} \\
    \midrule
    \multirow{3}{*}{ATB} & 
    \textbf{BERT} 
    & 188M
    & 82.99 & 82.99 & 82.99 & 18.87 
    & 188M
    & 82.96 & 83.17 & 83.07 & 19.09 \\
    & \ \  + SA 
    & 191M
    & \textbf{83.36} & 83.05 & 83.21 & 19.13 
    & 191M
    & 83.37 & 83.12 & 83.24 & 19.43 \\
    & \ \  + CatSA 
    & 192M
    & 83.33 & \textbf{83.20} & \textbf{83.27} & \textbf{20.04} 
    & 192M
    & \textbf{83.41} & \textbf{83.20} & \textbf{83.30} & \textbf{19.65} \\
    \midrule
    \multirow{6}{*}{CTB5} & 
    \textbf{BERT} 
    & 113M
    & 93.95 & 93.35 & 93.65 & 47.71 
    & 113M
    & 94.30 & 93.88 & 94.09 & 48.86 \\
    & \ \  + SA 
    & 116M
    & \textbf{94.07} & 93.39 & 93.73 & 49.43 
    & 116M
    & \textbf{94.80} & 93.73 & 94.26 & 49.14 \\
    & \ \  + CatSA 
    & 117M
    & 94.02 & \textbf{93.65} & \textbf{93.83} & \textbf{50.00} 
    & 117M
    & 94.70 & \textbf{94.00} & \textbf{94.35} & \textbf{50.00} \\
    \addlinespace[0.04cm]
    \cline{2-12}
    \addlinespace[0.04cm]
    & \textbf{ZEN} 
    & 235M
    & 93.82 & 93.65 & 93.73 & 50.29 
    & 235M
    & 94.37 & 93.69 & 94.03 & 48.87 \\
    & \ \  + SA 
    & 238M
    & 94.08 & 93.53 & 93.80 & 51.14  
    & 238M
    & 94.68 & 93.81 & 94.24 & 51.43 \\
    & \ \  + CatSA 
    & 239M
    & \textbf{94.23} & \textbf{93.66} & \textbf{93.94} & \textbf{51.41}  
    & 239M
    & \textbf{94.69} & \textbf{93.91} & \textbf{94.30} & \textbf{52.00} \\
    \midrule
    \multirow{9}{*}{PTB} & \textbf{BERT-LC} 
    & 344M
    & 95.71 & 95.53 & 95.62 & 54.06 
    & 344M
    & 95.71 & 95.61 & 95.66 & 53.35 \\
    & \ \  + SA 
    & 349M
    & 95.80 & \textbf{95.55} & 95.68 & 53.94 
    & 349M
    & 95.71 & 95.70 & 95.70 & 54.29 \\
    & \ \  + CatSA 
    & 350M
    & \textbf{96.02} & 95.51 & \textbf{95.77} & \textbf{54.64} 
    & 350M
    & \textbf{95.79} & \textbf{95.85} & \textbf{95.82} & \textbf{55.79} \\
    \addlinespace[0.04cm]
    \cline{2-12}
    \addlinespace[0.04cm]
    & \textbf{BERT-LU} 
    & 345M
    & 95.61 & 95.59 & 95.60 & 54.29 
    & 345M
    & 95.59 & 95.76 & 95.67 & 54.24 \\
    & \ \   + SA 
    & 350M
    & 95.61 & 95.71 & 95.66 & 54.24 
    & 350M
    & 95.69 & 95.75 & 95.72 & 54.53 \\
    & \ \   + CatSA 
    & 351M 
    & \textbf{95.76} & \textbf{95.74} & \textbf{95.75} & \textbf{55.29}
    & 351M
    & \textbf{95.77} & \textbf{95.84} & \textbf{95.80} & \textbf{54.71} \\
    \addlinespace[0.04cm]
    \cline{2-12}
    \addlinespace[0.04cm]
    & \textbf{XLNet-LC} 
    & 371M
    & 95.78 & 95.79 & 95.78 & 54.81 
    & 371M
    & 95.97 & 95.60 & 95.79 & 54.70 \\
    & \ \   + SA 
    & 375M
    & 95.83 & \textbf{95.95} & 95.89 & 54.94 
    & 375M
    & 95.92 & 95.95 & 95.93 & 55.71 \\
    & \ \   + CatSA 
    & 376M 
    & \textbf{96.02} & 95.84 & \textbf{95.93} & \textbf{55.88} 
    & 376M
    & \textbf{95.97} & \textbf{96.02} & \textbf{95.99} & \textbf{56.06} \\
    \bottomrule
\end{tabular}
\end{sc}
\end{small}
\end{center}
\vspace{-0.3cm}
\caption{
Experimental results in terms of precision (\textsc{P}), recall (\textsc{R}), F-score (\textsc{F1}) and complete match score (\textsc{M}) of our models on the development set of ATB, CTB5 and PTB with different configurations,
i.e., with and without POS, span attention (\textsc{SA}), and categorical span attention (\textsc{CatSA}).
The boldface is added to the highest result (\textsc{P}, \textsc{R}, \textsc{F1}, and \textsc{M}) within each group of three models (one from BERT/XLNet baseline, one with \textsc{SA}, and the other with \textsc{CatSA}).
For English, we use large cased (LC) version of BERT and XLNet and large uncased (LU) version of BERT.
\textsc{PARM} reports the number of trainable parameters in each model.}
\vskip -1.em
\label{tab: main results}
\end{table*}


\begin{table*}[t]
\begin{center}
\begin{small}
\begin{sc}
\begin{tabular}
    {L{6.18cm} |
    C{0.57cm} C{0.57cm} >{\columncolor[rgb]{0.95,0.95,0.95}}C{0.71cm} |
    C{0.57cm} C{0.57cm} >{\columncolor[rgb]{0.95,0.95,0.95}}C{0.71cm} |
    C{0.57cm} C{0.57cm} >{\columncolor[rgb]{0.95,0.95,0.95}}C{0.71cm}}
    \toprule
    \multirow{2}{*}{\textbf{Models}} & \multicolumn{3}{c|}{\textbf{ATB}} & \multicolumn{3}{c|}{\textbf{CTB5}} & \multicolumn{3}{c}{\textbf{PTB}} \\
    \cline{2-10}
    \addlinespace[0.04cm]
    & \textbf{P} & \textbf{R} & \textbf{F1} & \textbf{P} & \textbf{R} & \textbf{F1} & \textbf{P} & \textbf{R} & \textbf{F1} \\
    \midrule
    \newcite{green-manning-2010-better} 
    & 78.92 & 77.72 & 78.32
    & - & - & -  
    & - & - & - \\
    \newcite{shen-etal-2018-straight}
    & - & - & -
    & 86.6~~ & 86.4~~ & 86.5~~ 
    & 92.0~~ & 91.7~~ & 91.8~~ \\
    \newcite{teng-zhang-2018-two}
    & - & - & -
    & 88.0~~ & 86.6~~ & 87.3~~  
    & 92.5~~ & 92.2~~ & 92.4~~ \\
    \newcite{joshi-etal-2018-extending}
    & - & - & -
    & - & - & - 
    & 94.8~~ & 93.8~~ & 94.3~~ \\
    \newcite{suzuki-etal-2018-empirical}
    & - & - & -
    & - & - & - 
    & - & - & 94.32 \\
    \newcite{kitaev-klein-2018-constituency}
    & - & - & -
    & - & - & - 
    & 95.40 & 94.85 & 95.13  \\
    \newcite{kitaev-etal-2019-multilingual} (BERT)
    & - & - & -
    & 91.96 & 91.55 & 91.75  
    & 95.73 & 95.46 & 95.59 \\
    \newcite{fried-etal-2019-cross} (BERT)
    & - & - & -
    & - & - & 92.14 
    & - & - & 95.71 \\
    \newcite{zhou-zhao-2019-head} (BERT)
    & - & - & -
    & 92.03 & 92.33 & 92.18 
    & 95.70 & 95.98 & 95.84 \\
    \newcite{zhou-zhao-2019-head} (XLNet)
    & - & - & -
    & - & - & - 
    & 96.21 & 96.46 & 96.33 \\
    *\newcite{mrini2019rethinking} (BERT/XLNet + POS)
    & - & - & -
    & 91.85 & \textbf{93.45} & 92.64
    & 96.24 & \textbf{96.53} & 96.38 \\
    \midrule
    SCT \cite{manning-etal-2014-stanford}
    & 68.33 & 71.78 & 70.02 
    & $\dagger$ & $\dagger$ & $\dagger$ 
    & 86.21 & 86.73 & 86.47 \\
    BNP \cite{kitaev-klein-2018-constituency}
    & 72.84 & 76.59 & 74.67 
    & 91.83 & 91.53 & 91.68 
    & 95.46 & 94.89 & 95.17 \\
    \midrule
    BERT
    & 83.06 & 82.87 & 82.96
    & 92.16 & 91.98 & 92.07
    & 95.91 & 95.17 & 95.54 \\
    \ \  + SA
    & 83.25 & 82.85 & 83.05
    & 92.31 & 92.03 & 92.17
    & 96.04 & 95.40 & 95.72 \\
    \ \  + CatSA
    & 83.40 & \textbf{83.11} & 83.26 
    & 92.25 & 92.14 & 92.20
    & 96.11 & 95.58 & 95.85 \\
    ZEN/XLNet
    & - & - & -
    & 92.20 & 92.05 & 92.13
    & 96.52 & 95.70 & 96.11 \\
    \ \  + SA
    & - & - & -
    & 92.34 & 92.02 & 92.18
    & 96.58 & 96.03 & 96.31 \\
    \ \  + CatSA
    & - & - & - 
    & 92.50 & 91.98 & 92.24 
    & \textbf{96.64} & 96.07 & 96.36 \\
    \midrule
    *BERT + POS
    & 82.98 & 82.97 & 82.97
    & 92.52 & 92.06 & 92.29
    & 95.92 & 95.27 & 95.60 \\
    \ \ \  + SA
    & 83.36 & 82.80 & 83.08
    & 92.61 & 92.20 & 92.40
    & 95.96 & 95.51 & 95.73 \\
    \ \ \  + CatSA
    & \textbf{83.48} & 83.07 & \textbf{83.27}
    & \textbf{92.83} & 92.50 & \textbf{92.66} 
    & 96.09 & 95.62 & 95.86 \\
    *ZEN/XLNet + POS
    & - & - & -
    & 92.37 & 92.16 & 92.26
    & 96.42 & 95.86 & 96.14 \\
    \ \ \  + SA
    & - & - & -
    & 92.40 & 92.32 & 92.36
    & 96.56 & 96.10 & 96.33 \\
    \ \ \  + CatSA
    & - & - & - 
    & 92.61 & 92.42 & 92.52 
    & 96.61 & 96.19 & \textbf{96.40} \\
    \bottomrule
\end{tabular}
\end{sc}
\end{small}
\end{center}
\vspace{-0.4cm}
\caption{Comparing (in terms of  \textsc{P}recison, \textsc{R}ecall and \textsc{F1} scores) our best performing models (\textsc{BERT-LC} and \textsc{ZEN}/\textsc{XLNet-LC}) with previous studies and prevailing toolkits (i.e., SCT and BNP) on the test sets of ATB, CTB5 and PTB.
The results for SCT are not comparable to other systems including ours
(as indicated by $\dagger$) because
SCT is trained on a different dataset.
Models marked by * use predicted POS tags as additional input.
}
\label{tab: sota}
\vskip -1em
\end{table*}

\subsection{N-gram Lexicon Construction}

For n-gram extraction, 
we compute the pointwise mutual information (PMI) of any two adjacent words
$x', x''$ in the dataset by
\begin{equation}
\setlength\abovedisplayskip{5pt}
\setlength\belowdisplayskip{5pt}
    PMI(x', x'') = \log \frac{p(x'x'')}{p(x')p(x'')}
\end{equation}
where $p$ is the probability of an n-gram (i.e., $x'$, $x''$ and $x'x''$) in a dataset. 
A high PMI score suggests that the two words co-occur a lot in the dataset and are more likely to form an n-gram.
We set the threshold to 0 to determine whether a delimiter should be inserted between the two adjacent words $x'$ and $x''$.
%
In other words, to build the lexicon $\mathcal{N}$ from a dataset, we use PMI as an unsupervised segmentation method
to segment the dataset and collect all n-grams (n $\leq$ 5)\footnote{We empirically set the max n-gram length to 5 as a unified threshold for all three languages.} appearing at least twice in the training and development sets combined.\footnote{We show the details of extracting the lexicon with example n-grams in the Appendix.}

\subsection{Model Implementation}

In our experiments, we use BERT \cite{devlin-etal-2019-bert} as the basic encoder for all three languages 
%
and use ZEN \cite{Sinovation2019ZEN} and XLNet-large \cite{yang-2019-xlnet} for Chinese and English, respectively.%
\footnote{We download BERT models for Arabic and English from \url{https://github.com/google-research/bert}, and for Chinese from \url{https://s3.amazonaws.com/models.huggingface.co/}.
We download ZEN and XLNet at \url{https://github.com/sinovation/ZEN}
amd \url{https://github.com/zihangdai/xlnet}.}
For BERT, ZEN, and XLNet, we use the default hyper-parameter settings.
(e.g., 
24 layers with 1024 dimensional hidden vector for the large models).
In addition, following 
\newcite{kitaev-etal-2019-multilingual}, 
\newcite{zhou-zhao-2019-head} and \newcite{mrini2019rethinking}, we add three
additional token-level self-attention layers to the top of BERT, ZEN, and XLNet.

For other settings, we randomly initialize all n-gram embeddings used in our attention module\footnote{We also try initializing the n-grams with pre-trained embeddings \cite{pennington2014glove,song-etal-2018-directional,yamada2020wikipedia2vec}, where the results show small differences.}
with their dimension matching that of the hidden vectors obtained from the encoder (e.g., 1024 for BERT-large).
Besides, we run our experiments with and without predicted part-of-speech (POS) tags.
Following previous studies, for the experiments without POS tags, we take sentences as the only input;
for the experiments with POS tags, we obtain the POS tags from Stanford POS Tagger \cite{toutanova-etal-2003-feature} and incorporate the POS tags by directly concatenating their embeddings with the output of the BERT/ZEN/XLNet encoder.
%
%
%
%
%
%
%
Following previous studies \cite{suzuki-etal-2018-empirical,kitaev-etal-2019-multilingual}, we use hinge loss during the training process 
and evaluate different models by by precision, recall, F1 score, and complete match score via the standard evaluation toolkit \textsc{Evalb}\footnote{\url{https://nlp.cs.nyu.edu/evalb/}}.

During the training process, we try three learning rates, i.e., 5e-5, 1e-5, 5e-6, with a fixed random seed, pick the model with the best F1 score on the development set, and evaluate it on the test set.


\section{Results and Analyses}

\subsection{Overall Performance}


In the main experiment, we compare the proposed models with and without the span attention to explore the effect of the span attention on chart-based constituency parsing.
For models with the span attention, we also run the settings with and without the categorical mechanism.
The results (i.e., precision, recall, F1 score, and complete match scores of all models, as well as their number of trainable parameters) 
with different configurations (including whether to use the predicted POS tags) on the development sets of ATB, CTB5, and PTB are reported in Table \ref{tab: main results}.

There are several observations.
First, the span attention over n-grams shows its generalization ability, where 
consistent improvements of F1 over the baseline models are observed on all languages under different settings (i.e., with and without using predicted POS tags; using BERT or XLNet encoders).
%
%
%
Second, compared with span attention without the category mechanism, in which n-grams are weighted together, 
models with categorical span attention perform better on both F1 and complete match scores with a relatively small increase of parameter numbers (around $1M$).
Particularly, for the complete match scores, the span attention with normal attentions does not outperform the baseline models in some cases, whereas the categorical span attention mechanism does in all cases.
These results could be explained by that frequent short n-grams dominate the general attentions so that the long ones containing more contextual information fail to function well in filling the missing information in the span representation, and thus harm the understanding of long spans, which results in inferior results in complete match score.
In contrast, the categorical span attention is able to weight n-grams in different length separately, so that the attentions are not dominated by high-frequency short n-grams and thus reasonable weights can be assigned to long n-grams.
Therefore, our model can learn from the important long n-grams and have a good performance on the long spans, which results in consistent improvements over baseline models in complete match scores.
Third, on CTB5,
%
models with ZEN encoder consistently outperform the ones with BERT without using POS tags, while they fail to do so with the POS tags as the additional input, which suggests that the predicted POS tags may have more conflict with ZEN compared with BERT.

\begin{table}[t]
\begin{center}
\begin{small}
\begin{sc}
\begin{tabular}{l | R{1.0cm} R{0.9cm}}
    \toprule
    \textbf{Models} & \textbf{Brown} & \textbf{Genia} \\
    \midrule
    BERT \cite{fried-etal-2019-cross}
    & 93.10 & 87.54 \\
    \midrule
    BERT 
    & 93.13 & \textbf{87.58} \\
    \ \  + SA 
    & 93.24 & 87.50 \\
    \ \  + CatSA 
    & \textbf{93.29} & 87.53 \\
    \bottomrule
\end{tabular}
\end{sc}
\end{small}
\end{center}
\vspace{-0.3cm}
\caption{
Cross-domain experiment results (\textsc{F1} scores) from previous studies
and our models (based on \textsc{BERT-LC}), on the entire Brown and Genia corpora when trained from the training set of PTB.}
\label{tab: cross-domain results}
\vskip -1em
\end{table}

Moreover, we run our models on the test set of each dataset and compare the results with previous studies, as well as the ones from prevailing parsers, i.e., Stanford CoreNLP Toolkits (SCT)\footnote{We use the version of 3.9.2 obtained from \url{https://stanfordnlp.github.io/CoreNLP/}.} \cite{manning-etal-2014-stanford} and Berkeley Neural Parser (BNP)\footnote{We obtain their models from \url{https://github.com/nikitakit/self-attentive-parser}.} \cite{kitaev-klein-2018-constituency}.
The results are reported in Table \ref{tab: sota}, where the models using predicted POS tags are marked with ``*''.\footnote{For our models with BERT encoder, we only report the results of the ones using the cased version of BERT-large.}
Our models with \textsc{CatSA} outperform previous best performing models from \citet{zhou-zhao-2019-head} and \citet{mrini2019rethinking} under different settings (i.e., whether to use the predicted POS tags), and achieve state-of-the-art performance on all datasets.
Compared with \citet{zhou-zhao-2019-head} and \citet{mrini2019rethinking} which improve constituency parsing by leveraging the dependency information when training their head phrase structure grammar (HPSG) parser, our approach enhances the task from another direction by incorporating n-gram information through the span attentions as a way to address the limitation of using hidden vector subtraction to represent spans.

\subsection{Cross-domain Experiments}

To further explore whether our approach can be generalized across domains, we follow the setting of \citet{fried-etal-2019-cross} to conduct cross-domain experiments on the Brown and Genia corpus using the models with \textsc{SA} and \textsc{CatSA}, as well as their corresponding baseline.
Note that, for fair comparison, we use BERT-large cased as the encoder without using the predicted POS tags.
We follow \citet{fried-etal-2019-cross} to train models on the training set of PTB and evaluate them on the entire Brown corpus and the entire Genia corpus.
To construct $\mathcal{N}$ in this experiment,
we extract n-grams by PMI from the training set of PTB.
The results (F1 scores) are reported in Table \ref{tab: cross-domain results}.
%
From the table, we find that our model with categorical span attentions (+ \textsc{CatSA}) outperforms the BERT baseline \cite{fried-etal-2019-cross} on the Brown corpus while fails to do so on the Genia corpus.
The explanation cloud be that the distance between Genia (medical domain) and PTB (news wire domain) is much larger than that between Brown and PTB,
so that the n-gram overlap in two domains are limited and
thus has little influence to the target domain.

\subsection{Effect of \textsc{CatSA} on Long Sentences}

\begin{figure}[t]
    \centering
    \includegraphics[width=0.48\textwidth, trim=0 20 0 0]{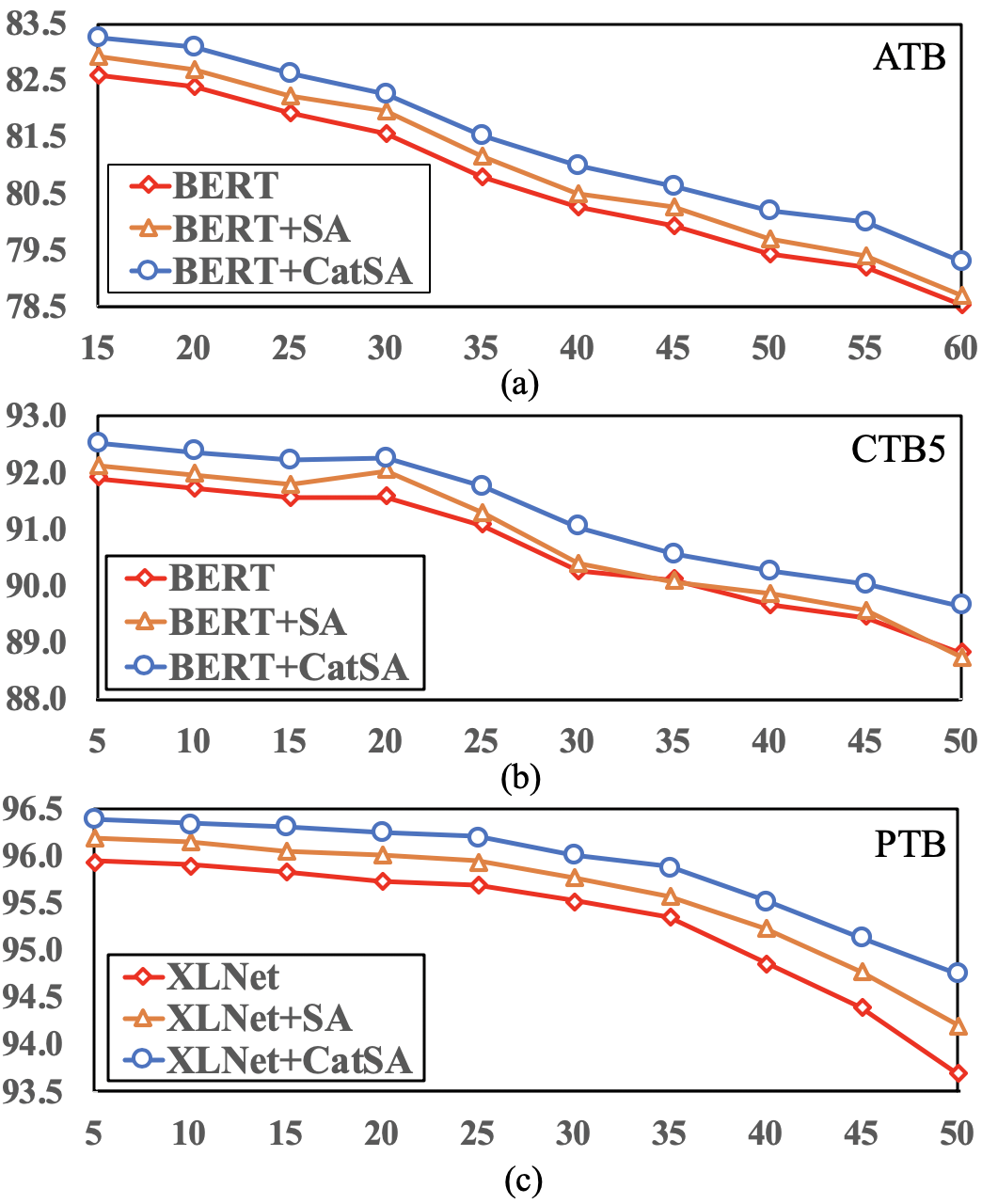}
    \caption{The F1 curves with respect to the minimal test sentence length (the horizontal axis) of different models performed on ATB (a), CTB (b), and PTB (c).
    }
    \label{fig: sent_len}
    \vskip -1em
\end{figure}

To explore the effect of our approach, we investigate our best performing models (where predicted POS tags are used) with the span attention module and the corresponding baselines on different length of sentences in the test sets.
The curves of F1 scores with respect to the minimal test sentence length (the horizontal axis) from different models on ATB, CTB5, and PTB are illustrated in Figure \ref{fig: sent_len}(a), \ref{fig: sent_len}(b), and \ref{fig: sent_len}(c), respectively.\footnote{Given the variance of average sentence length in different datasets (see Table \ref{tab: dataset details}), we set the minimal length from 5 to 50 on CTB5 and PTB, and 15 to 60 on ATB, with a step of 5.}

In general, long sentences are harder to parse and thus all models' performance degrades when sentence length increases.
Yet, our models with CatSA outperform the baseline for all sentence groups
and the gap is bigger for long sentences, which indicates our approach can handle long sentences better than the baselines.
%
%
One possible explanation for this is that long sentences will have larger text spans and may require more long-distance contextual information.
Our approach incorporates n-gram information into the span representation and thus can appropriately leverages the infrequent long n-grams by separately weighting them in different categories.

\begin{figure}[t]
    \centering
    \includegraphics[width=0.48\textwidth, trim=0 15 0 0]{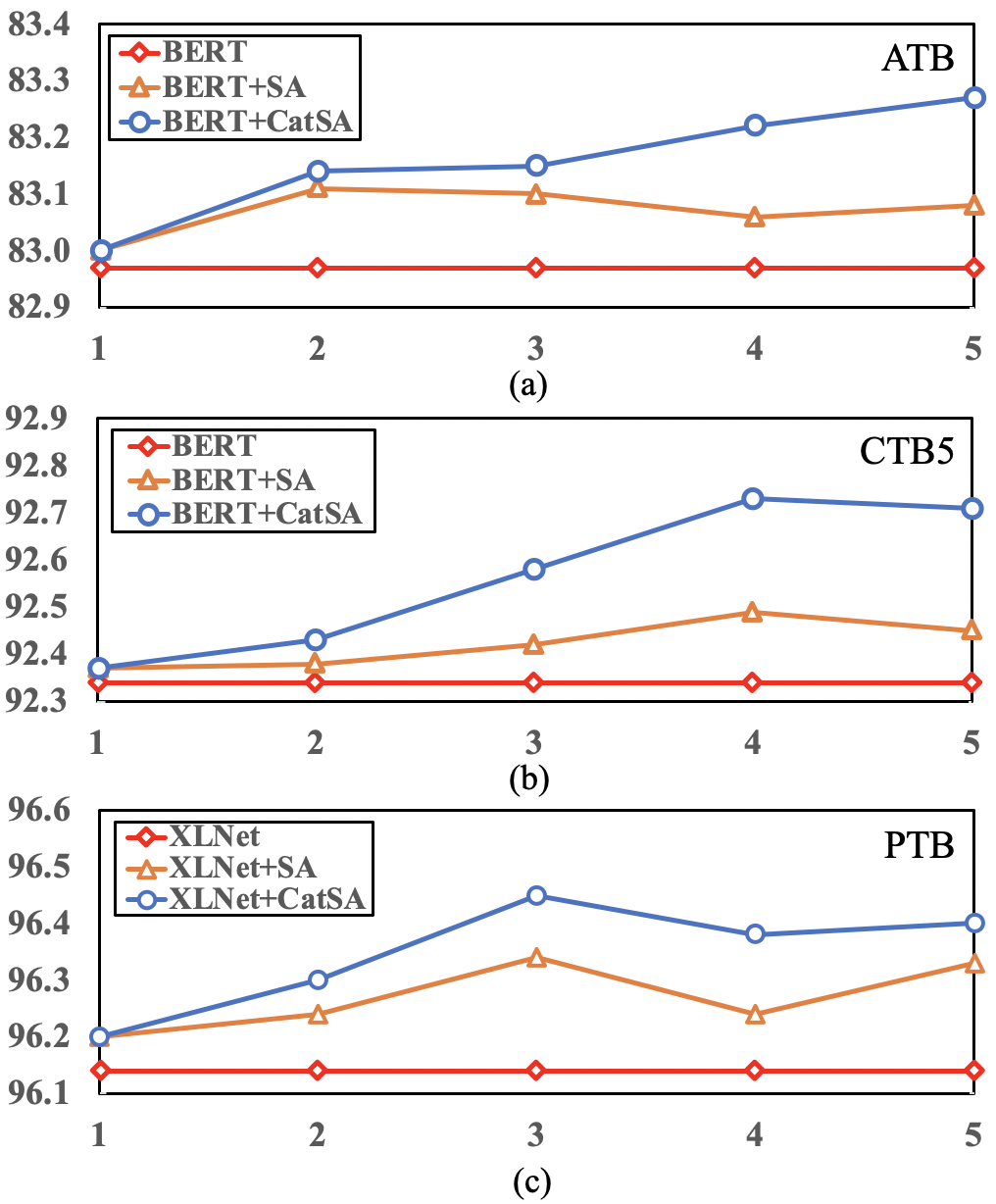}
    \caption{The F1 curves with respect to the max length (the horizontal axis) of n-grams used in different models performed on ATB (a), CTB (b), and PTB (c).
    }
    \label{fig: n-gram_len}
    \vskip -1em
\end{figure}

\begin{figure}[t]
    \centering
    \includegraphics[width=0.48\textwidth, trim=0 15 0 0]{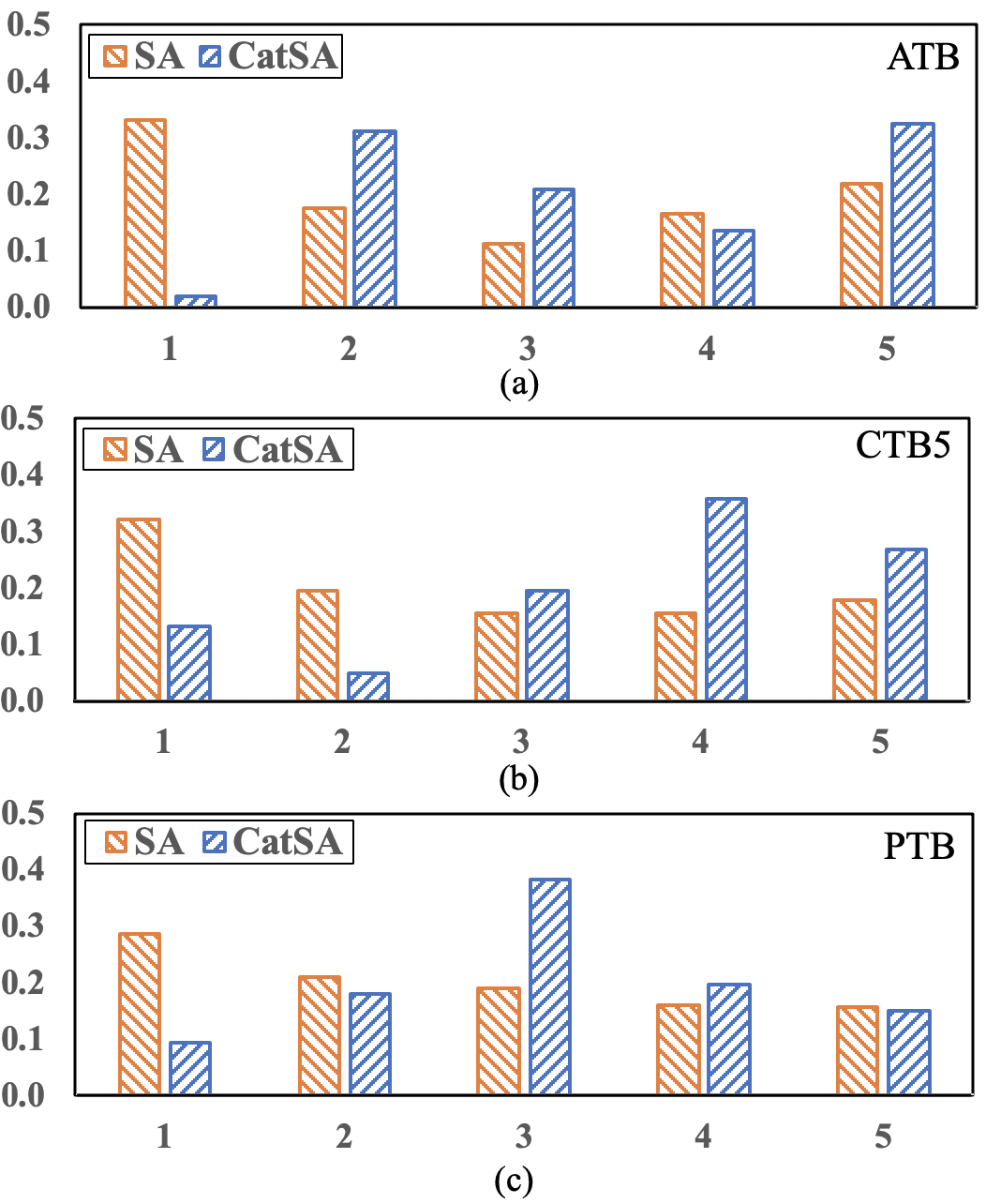}
    \caption{The histograms of average weights assigned to 
    n-gram categories
    in different lengths,
    with weights from 
    \textsc{SA} and 
    \textsc{CatSA} 
    show different patterns.
    }
    \label{fig: ngram_len}
    \vskip -1em
\end{figure}

\subsection{Analysis on Different N-gram Lengths}

To test using n-grams in different length, we conduct an ablation study on the n-grams with respect to their length.
In doing so, we conduct experiments on the best performing models (where predicted POS tags are used) with the span attention module, by restricting that n-grams whose length are larger than a threshold is excluded from the lexicon $\mathcal{N}$.
We try the threshold from 1 to 5 and demonstrate the curves (F-scores) on the test set of ATB, CTB5, and PTB in Figure \ref{fig: n-gram_len}(a), (b), and (c), respectively.
The results of their corresponding baselines are also represented in red curves for reference.
It is found from the curves that our models with span attentions consistently outperform the baseline models, which indicates the robustness of our approach with respect to different n-grams used in the model.
In addition, for different languages, the n-gram threshold varies when the best performance is obtained.
For example, the best performing model on English is with three words as the maximum length of n-grams, while that
is five for Arabic and four for Chinese.

Moreover, 
to investigate how the categorical span attention addresses the problem that high-frequency short n-grams can dominate the general attentions, we run the best performing models
with span attentions on the whole ATB, CTB5, and PTB datasets, obtain the total weight assigned to each n-gram, and compute the average weight for the n-grams in each n-gram length category.
Figure \ref{fig: ngram_len} shows the histograms of the average weights from models with \textsc{SA} and \textsc{CatSA}.

The histograms show that the models with \textsc{SA} (the orange bars) tend to assign short n-grams relatively high weights, especially the uni-grams.
This is not surprising because short n-grams occur more frequently and are thus updated more times than long ones.
In contrast, the models with \textsc{CatSA} show a different weight distribution (the blue bars) among n-grams with different lengths, which indicates that the \textsc{CatSA} module could balance the weights distribution and thus enable the model to learn from infrequent long n-grams.
%

\subsection{Case Study}


\begin{figure}[t]
    \centering
    \includegraphics[width=0.48\textwidth, trim=0 20 0 -5]{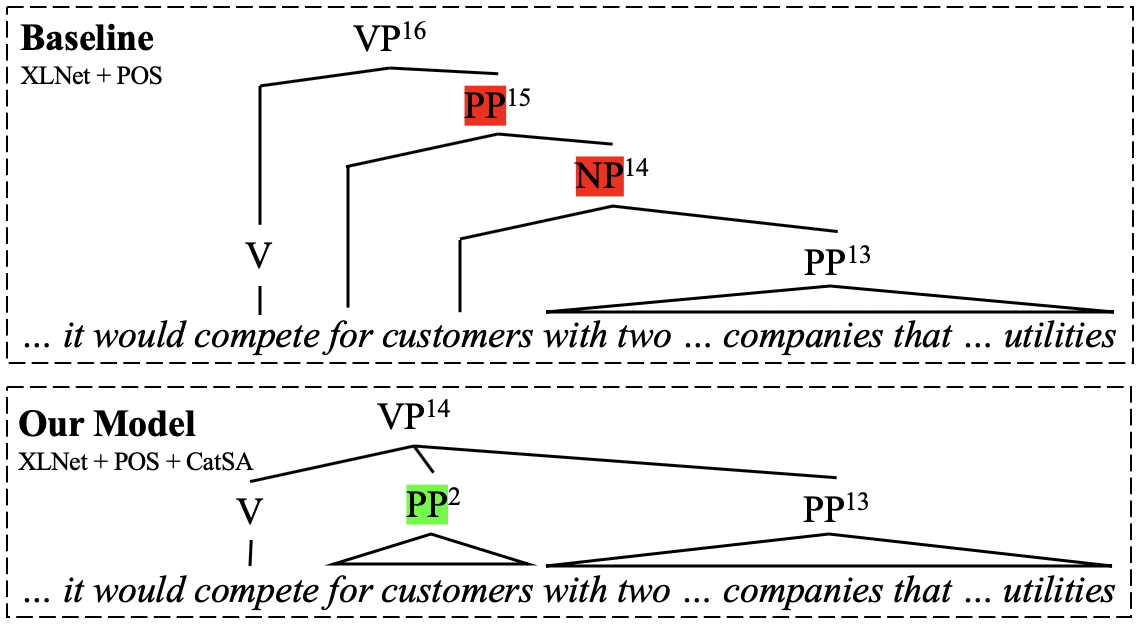}
    \caption{
    An example sentence with its parsing results from the best performing baseline and our model.
    The correct and wrong parsing results are highlighted on the span labels by green and red, respectively.
    The superscripts on the span labels illustrate the heights of them. ``V'' is a POS tag so there is no height for it.
    }
    \label{fig:case_study}
    \vskip -1em
\end{figure}

To illustrate how our model improves baselines with the span attention, especially for long sentences, we show the parse trees produced by the two models for an example sentence in Figure \ref{fig:case_study}, where the superscript for the internal node is the height of the subtree rooted at that node.
%
%
%
%
%
%
%
In this case, our model correctly attaches the ``\textit{PP}'' (``\textit{with two ... utilities}'') containing 24 words to the verb ``\textit{compete}'', while the baseline attach it to the noun ``\textit{customers}''.
Since the distances between the boundary positions of the wrongly predicted spans (highlighted in red) are relatively long, the baseline system, which simply represents the span as subtraction of the hidden vectors at the boundary positions, may fail to capture the important context information within the text span.   
In contrast, the span representations used in our model are enhanced by weighted n-gram information and thus contain more context information.
Therefore, in deciding which component (i.e., ``\textit{compete}'' or ``\textit{customer}'')  the \textit{with}-PP 
should attach to, n-grams (e.g., the uni-gram ``\textit{companies}'') may provide useful cues, since "customers with companies" is less likely than ``compete with companies".

\section{Related Work}

There are two main types of parsing methodologies.
%
One is the transition-based approaches \cite{sagae-lavie-2005-classifier};
the other is the chart-based approaches \cite{collins-1997-three,glaysher-moldovan-2006-speeding}.
Recently, neural methods start to play a dominant role in this task, where 
improvements mainly come from powerful encodings \cite{dyer-etal-2016-recurrent,cross-huang-2016-span,liu-zhang-2017-order,stern-etal-2017-minimal,gaddy-etal-2018-whats,kitaev-klein-2018-constituency,kitaev-etal-2019-multilingual,fried-etal-2019-cross}.
Moreover, there are studies that do not follow the aforementioned methodologies, which instead 
regard the task as a sequence-to-sequence generation task \cite{NIPS2015_5635,suzuki-etal-2018-empirical}, a language modeling \cite{choe-charniak-2016-parsing} task or a sequence labeling task \cite{gomez-rodriguez-vilares-2018-constituent}.
%
To further improve the performance, some studies leverage extra resources (such as auto-parsed large corpus \cite{NIPS2015_5635}, pre-trained word embeddings \cite{kitaev-klein-2018-constituency}), HPSG information \cite{zhou-zhao-2019-head, mrini2019rethinking},
or use model ensembles \cite{kitaev-etal-2019-multilingual}.
%
%
Compared to these studies, our approach offers an alternative way to enhance constituency parsing with effective leveraging of n-gram information.
%
Moreover, the proposed span attention addresses the limitation of previous studies \cite{kitaev-klein-2018-constituency,kitaev-etal-2019-multilingual} that spans are represented by the subtraction of encoded vectors at span boundaries (i.e., the hidden states at initial and ending positions of the span) and thus reduces information loss accordingly.
In addition, the categorical span attention provides a simple, yet effective, improvement over the normal attention to process n-grams in a more precise way,
which could become a reference for leveraging similar resources in future research.

\vspace{-0.1cm}
\section{Conclusion}
\vspace{-0.1cm}

In this paper, we proposed span attention to integrate n-gram into span representations to enhance chart-based neural constituency parsing.
Specifically, for each text span in an input sentence, we firstly extracted n-grams in that span from an n-gram lexicon, and then fed them into the span attention to weight them according to their contribution to the parsing process.
To better leverage n-grams, especially the long ones, categorical span attention was proposed to improve the normal attention by categorizing n-grams according to their length and weighting them separately within each category.
Such span attention not only leverages important contextual information from n-grams but also addresses the limitation of current Transformer-based encoders using subtraction for span representations.
To the best our knowledge, this is the first work using n-grams for neural constituency parsing.
%
The effectiveness of our approach was demonstrated by experimental results on three benchmark datasets from 
Arabic, Chinese, and English, where 
state-of-the-art performance is obtained on all of them.
%

\bibliography{emnlp2020}
\bibliographystyle{acl_natbib}

\appendix


\vspace{1cm}

\section*{Appendix A: Extracting the Lexicon using PMI} \label{app: pmi}

In our experiments, we build the n-gram lexicon $\mathcal{N}$ based on pointwise mutual information (PMI)
with n-gram probability estimated from the union of the training and development set of each dataset.
Specifically, we compute the PMI of any two adjacent words $x', x''$ in the dataset by
\begin{equation}
\setlength\abovedisplayskip{5pt}
\setlength\belowdisplayskip{5pt}
    PMI(x', x'') = log \frac{p(x'x'')}{p(x')p(x'')}
\end{equation}
where $p$ is the probability of an n-gram (i.e., $x'$, $x''$ and $x'x''$) in the dataset.
A high PMI score suggests that the two words co-occur frequently in the dataset and are more likely to form an n-gram.
For each pair of adjacent words $x_{i-1}$, $x_{i}$ in a sentence $\mathcal{X}=x_1x_2 \cdots x_{i-1}x_{i} \cdots x_{n}$, we use a threshold to determine whether a delimiter should be inserted in between them.
As a result, the sentence $\mathcal{X}$ is segmented into pieces of n-grams; we extract those n-grams to form the lexicon $\mathcal{N}$.
For example, for a given sentence
\begin{equation}
\setlength\abovedisplayskip{5pt}
\setlength\belowdisplayskip{5pt}
    \mathcal{X}=x_1 x_2 x_3 x_4 x_5
\end{equation}
and the PMI of all adjacent words (i.e., $x_1 x_2$, $x_2 x_3$, $x_3 x_4$, $x_4 x_5$) in it, where
\begin{align}
\setlength\abovedisplayskip{5pt}
\setlength\belowdisplayskip{5pt}
    PMI(x_2, x_3) > t \\
    PMI(x_3, x_4) > t
\end{align}
and 
\begin{align}
\setlength\abovedisplayskip{5pt}
\setlength\belowdisplayskip{5pt}
    PMI(x_1, x_2) < t \\
    PMI(x_4, x_5) < t
\end{align}
with $t$ denoting the threshold.
We add delimiters (denoted by ``/'') between $x_1$ and $x_2$, and $x_4$ and $x_5$ since their PMI is lower than $t$.
As a result, we obtain a segmented sentence
\begin{equation}
\setlength\abovedisplayskip{5pt}
\setlength\belowdisplayskip{5pt}
    \mathcal{X}'=x_1 / x_2 x_3 x_4 / x_5
\end{equation}
and from which we are able to extract three n-grams, i.e., $x_1$, $x_2 x_3 x_4$, and $x_5$ accordingly.

\begin{table}[t]
\begin{center}
\begin{tabular}{l | c }
    \toprule
    \textsc{N-grams} & \textsc{Avg. Attention} \\
    \midrule
    \textit{said} & 0.0141 \\
    \textit{more than} & 0.0063 \\
    \textit{as well as} & 0.0167 \\
    \textit{from a year earlier} & 0.0267 \\
    \textit{in the past few years} & 0.0184 \\
    \bottomrule
\end{tabular}
\vspace{-0.3cm}
\end{center}
\caption{Example n-grams with their average weights obtained from our best performing model (i.e., XLNet + POS + \textsc{CatSA}) on the entire PTB dataset.}
\label{tab: n-gram example}
\vskip -1em
\end{table}

\section*{Appendix B: N-gram Examples in the Lexicon $\mathcal{N}$}

To explore the effect of each individual n-gram in the lexicon $\mathcal{N}$,
we rank the n-grams according to their contributions to the constituency parsing task.
In doing so, we firstly run our best performing models (BERT/XLNet encoders with predicted POS tags) with categorical span attentions (+ \textsc{CatSA}) for Arabic, Chinese, and English on the entire ATB, CTB5, and PTB datasets, respectively.
Then, for each n-gram, we compute its average attention weights
according to its appearance in the entire dataset.
%
Afterwards, we group n-grams by their length and rank the n-grams according to their average attention weights within each group.
The top 50 n-grams in each group as well as their attention weights for each language are reported in the supplemental material.
As a demonstration, Table \ref{tab: n-gram example} shows a few n-grams with their average attention weights on the entire PTB dataset.

\end{document}